\documentclass[preprint,12pt]{elsarticle}
\usepackage[utf8]{inputenc}
\usepackage{times}
\usepackage{amsthm,amsmath,amsfonts,latexsym,amssymb}
\usepackage{graphics,fancyhdr,graphicx}
\usepackage{multirow}
\usepackage{float}
\usepackage{fleqn}
\usepackage{subfigure}
\usepackage{verbatim}
\usepackage{rotating}
\usepackage{makecell}
\usepackage{bm}
\usepackage{color}
\usepackage{bm}

\usepackage{colortbl}
\usepackage[table]{xcolor}

\usepackage{geometry}
\usepackage{mathrsfs}
\biboptions{numbers,sort&compress}
\definecolor{listinggray}{gray}{0.9}
\definecolor{lbcolor}{rgb}{0.9,0.9,0.9}
\definecolor{Darkgreen}{RGB}{0,100,0}

\begin{document}
\abovedisplayskip=6.0pt
\belowdisplayskip=6.0pt
\begin{frontmatter}

\title{LNO: Laplace Neural Operator for Solving Differential Equations}
\author[1,2]{Qianying Cao}
\ead{qianying_cao@brown.edu}
\author[2]{Somdatta Goswami}
\ead{somdatta_goswami@brown.edu}
\author[2,3]{George Em Karniadakis\corref{cor1}}
\ead{george_karniadakis@brown.edu}

\address[1]{State Key Lab of Coastal and Offshore Engineering, Dalian University of Technology}
\address[2]{Division of Applied Mathematics, Brown University}
\address[3]{School of Engineering, Brown University}

\cortext[cor1]{Corresponding author.}

\begin{abstract}
We introduce the Laplace neural operator (LNO), which leverages the Laplace transform to decompose the input space. Unlike the Fourier Neural Operator (FNO), LNO can handle non-periodic signals, account for transient responses, and exhibit exponential convergence. LNO  incorporates the pole-residue relationship between the input and the output space, enabling greater interpretability and improved generalization ability. Herein, we demonstrate the superior approximation accuracy of a single Laplace layer in LNO over four Fourier modules in FNO in approximating the solutions of three ODEs (Duffing oscillator, driven gravity pendulum, and Lorenz system) and three PDEs (Euler-Bernoulli beam, diffusion equation, and reaction-diffusion system). Notably, LNO outperforms FNO in capturing transient responses in undamped scenarios. For the linear Euler-Bernoulli beam and diffusion equation, LNO's exact representation of the pole-residue formulation yields significantly better results than FNO. For the nonlinear reaction-diffusion system, LNO's errors are smaller than those of FNO, demonstrating the effectiveness of using system poles and residues as network parameters for operator learning. Overall, our results suggest that LNO represents a promising new approach for learning neural operators that map functions between infinite-dimensional spaces.
\end{abstract}

\end{frontmatter}

\section{Introduction} 
\label{sec:introduction}

Real-world problems in computational science and engineering that lack closed-form solutions often require the use of expensive numerical solvers taxing CPU/GPU resources substantially both in processing and memory. Even minor changes to the problem's parameters frequently require running the numerical solver again, adding to the computational expense and time. However, the recent advancements in scientific machine learning, particularly the development of deep neural operators, offer a promising approach to solving parametrized ordinary/partial differential equations (ODEs/PDEs) using data-driven supervised learning. This approach offers an alternative and often advantageous alternative to traditional numerical solvers. By training a deep neural operator on a sufficiently large and diverse dataset offline, it is possible to approximate the solution of an ODE/PDE over a wide range of parameter values very fast online without further training.

Two neural operators that have shown promising results in approximating complex physical processes are the deep operator network introduced in $2019$ \cite{lu2021learning} and the Fourier neural operator (FNO) introduced in $2020$ \cite{li2020fourier}. In this work, we are particularly interested in developing a new neural operator, which addresses a specific bottleneck of FNO. Let us first recall that FNO is based on replacing the kernel integral operator with a convolution operator defined in Fourier space by employing a fast Fourier transform on the input space. The Fourier transform converts a physical system from the time domain to the frequency domain. From the definition of the Fourier transform, we have that the Fourier transform of a time-domain function $\boldsymbol x(t)$ is a continuous sum of exponential functions of the form $e^{-i\omega t}$, which means it adds the waves of positive and negative frequencies, where $\omega$ is the frequency. However, there are functions for which the Fourier transform does not exist such as $|\boldsymbol x(t)|$ because it is not absolutely integrable. Also, if we are interested in analyzing unstable systems then the Fourier transform cannot be used. The Laplace transform, on the other hand, redefines the Fourier transform and includes an exponential convergence factor $\sigma$ along with $i\omega$. Therefore, using the Laplace transform, the time-domain signal $\boldsymbol x(t)$ can be represented as a sum of complex exponential functions of the form $e^{-st}$, where $s = \sigma +i\omega$.

\noindent The Fourier transform, $\mathscr{F}$ of $\boldsymbol x(t)$, and its first and second derivatives can be defined as:
\begin{align}\label{eq:fourier_transform}
    \mathscr{F}\{\boldsymbol x(t)\} =\int_{t=-\infty}^{\infty}\boldsymbol x(t)e^{-i\omega t}\;dt, \;\;\; 
    \mathscr{F}( \dot{\boldsymbol x}(t)) = i\omega \mathscr{F}\{\boldsymbol x(t)\},\;\;\;
    \mathscr{F}( \ddot{\boldsymbol x}(t)) = -\omega^2\mathscr{F}\{\boldsymbol x(t)\},
\end{align}
which shows that the Fourier transform is not a suitable candidate for learning solutions for different initial conditions, as the transformation has no term to take the initial value into account. On the other hand, the Laplace transform, $\mathscr{L}$ of $\boldsymbol x(t)$ can be defined as:
\begin{align}
    \mathscr{L}\{\boldsymbol x(t)\}=\int_{t=0}^{\infty}\boldsymbol x(t)e^{-st}\;dt,
\end{align}
which is precisely motivated by the property that the differentiation of the time-dependent function with respect to time corresponds to the multiplication of the transform $\mathscr{L}(s)$ by $s$~\cite{kreyszig2010advanced}. More precisely,
\begin{align}
    \mathscr{L}( \dot{\boldsymbol x}) &= \mathscr{L}\{\boldsymbol x(t)\} - \boldsymbol x(0),\\
    \mathscr{L}( \ddot{\boldsymbol x}) &= s^2\mathscr{L}\{\boldsymbol x(t)\} - s\boldsymbol x(0) - \dot{\boldsymbol x}(0),  
\end{align}
which clearly depicts that the transformation takes into account the initial conditions (displacement and velocity), hence making it an appropriate candidate for learning neural operators with transient responses caused by zero initial conditions. 

\begin{figure}[ht]
\centering
\includegraphics[width=\textwidth, trim = 0cm 2cm 0cm 0cm, clip]{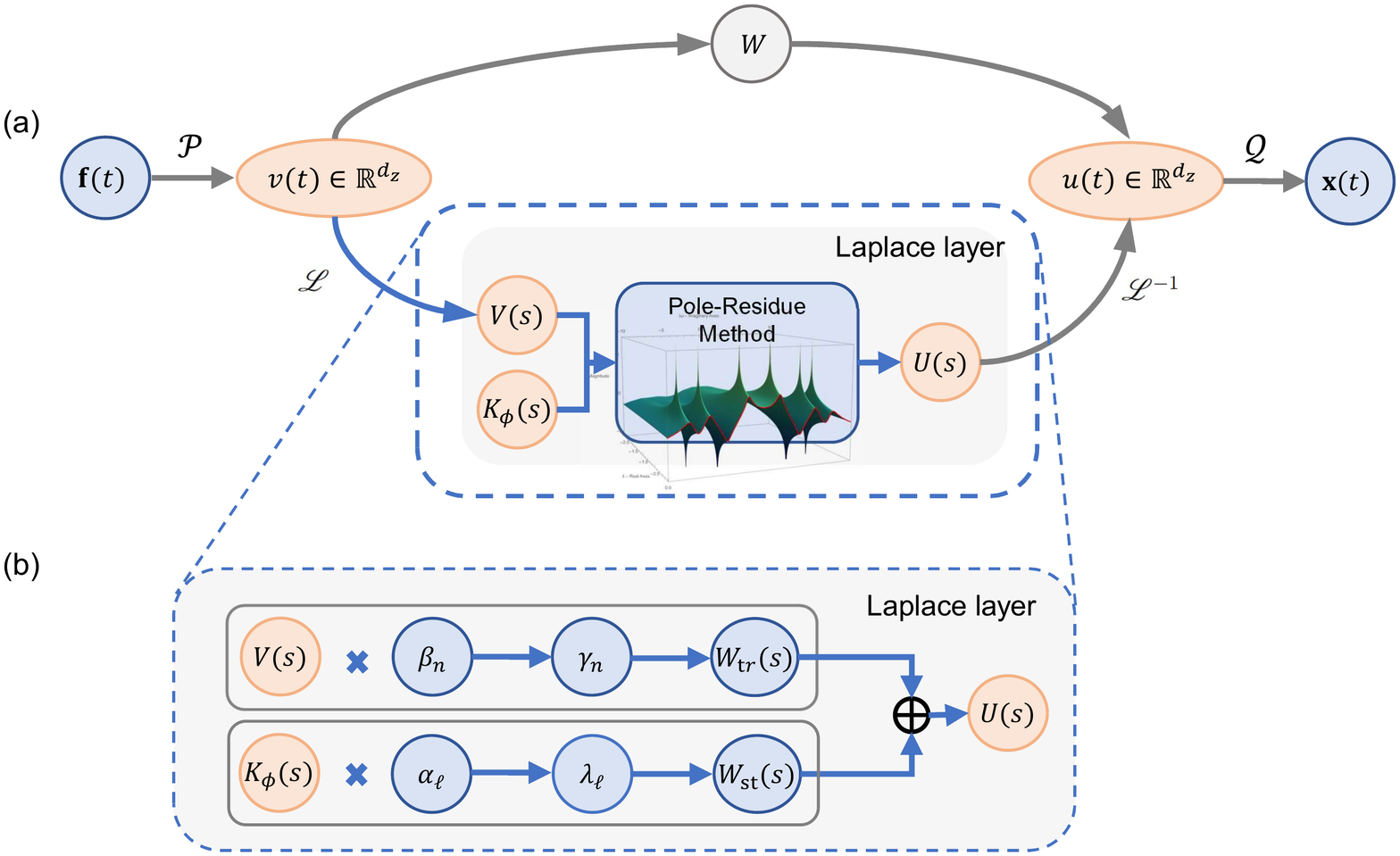}\\
\caption{(a) Schematic representation of the full architecture of Laplace neural operator (LNO). We start from an input function $\mathbf f(t)$ and follow the following steps. $1.$ Lift the input function to a higher dimension by a shallow neural network $\mathcal{P}$. $2.$ Apply a Laplace layer and a local linear transform $W$. $3.$ Project the output, $u(t)$, back to the target dimension employing a shallow neural network, $\mathcal{Q}$. $(b)$ Laplace layer: start from input $V(s)$. Top row: apply the pole-residue method to compute the transient response residues $\gamma_n$ based on system poles $\mu_n$ and residues $\beta_n$; express the transient response in the Laplace domain. Bottom row: apply the pole-residue method to compute the steady-state response residues $\lambda_{\ell}$ based on input poles $i\omega_{\ell}$ and residues $\alpha_{\ell}$; express the steady-state response in Laplace domain.}\label{Figure_neural_operator}
\end{figure}

Motivated by the limitations of FNO, especially for capturing the transient responses and non-period signals, and in order to exploit the advantages of the Laplace transformation, we propose the Laplace neural operator (LNO), that considers the decomposition of the input space employing the Laplace transform. The main idea behind this work is to employ a Laplace layer to replace the multiple Fourier layers of FNO so that the network parameters (including the system poles $\mu_n$ and residues $\beta_n$) are learned in the Laplace domain. The Laplace layer learns transient and steady-state responses simultaneously, as opposed to the Fourier layer which is more suited to learn the steady-periodic response. It is important to note that this work is not a trivial extension of replacing the Fourier modules of FNO with a Laplace layer but the approach also provides a more meaningful and physically interpretable mapping between the input and the output space on the Laplace domain by employing the poles and residue formulation. A schematic of the proposed neural operator is presented in Fig.~\ref{Figure_neural_operator}. 

The main contributions of this work can be summarized as follows:
\begin{itemize}
\item A novel framework to perform operator learning in the Laplace domain is proposed for solving ordinary and partial differential equations.
\item The physically meaningful pole-residue relationship between the functions in the input space and the system response is introduced into the network, which makes the operator more interpretable and endows it with good generalization ability.
\item The proposed framework can learn both transient and steady-periodic responses and, therefore, can be especially employed for systems without damping. 
\end{itemize}

The remainder of the paper is organized as follows. In Section \ref{sec:laplace_operator}, we introduce the Laplace neural operator. In Section \ref{sec:results}, we compare the accuracy of the proposed LNO with FNO and gated recurrent units (GRU) \cite{vlachas2019forecasting} for three time-dependent ODEs with transient responses. Finally, we summarize our observations and provide concluding remarks in Section \ref{sec:summary_and_discussions}.

\section{Neural Operators}
\label{sec:laplace_operator}

Neural operators are powerful machine learning models that can learn nonlinear mappings between infinite-dimensional functional spaces on bounded domains. They offer a unique simulation framework for predicting multi-dimensional complex dynamics in real-time. Once trained, the neural operators are discretization invariant, meaning they can be applied across different parameterizations of the underlying functional data without requiring retraining. This makes them highly versatile tools for being employed as an efficient surrogate model for learning parametrized ODEs/PDEs that govern physical systems. Inspired by the universal approximation theorem of operators proposed by Chen \& Chen \cite{chen1995universal}, the first neural operator, deep operator network (DeepONet) was proposed in $2019$~\cite{lu2021learning}. DeepONet is represented by a summation of the products of two or more deep neural networks (DNNs), corresponding to the branch NN/s for the input function/s and the trunk NN for the output function. All the NNs have general architectures, e.g., the branch NN can be replaced with a CNN or a ResNet, and the trunk NN can be a fully connected NN or could be a network with proper orthogonal modes etc. Motivated by the idea of neural operators, the Fourier neural operator (FNO) was proposed in $2020$, which employs the Green's function as its backbone and the convolution integral kernel in the Green's function is parameterized directly in the Fourier space~\cite{li2020fourier}.

As discussed in Section~\ref{sec:introduction}, herein we  introduce the Laplace neural operator (LNO), which could improve the approximation capacity of FNO in cases of transient responses and no damping conditions for time-dependent ODEs/PDEs. Let $\Omega$ be a bounded open set in $\mathbb{R}^D$, and let $\mathcal{X}$ and $\mathcal{Y}$ be separable Banach spaces defined on $\Omega$ with dimensions $d_x$ and $d_y$, respectively. Suppose that a non-linear map $\mathcal{G}:\mathcal{X} \rightarrow \mathcal{Y}$ arises from the solution of a time-dependent PDE. The goal is to approximate this non-linear operator by a parametric mapping as $\mathcal{G}_{\theta}: \mathcal{X} \rightarrow \mathcal{Y}$, where $\theta$ are the network parameters that are learned through backpropagation, while training of the neural operator, based on the labeled input-output pairs $\{\mathbf{f}_j, \mathbf{u}_j\}_{j=1}^N$ generated on a discretized domain $\Omega$, where $f_j\sim u_j$ is an i.i.d sequence from the probability measure $\mathbf u$ supported on $\mathcal{X}$. 

\subsection{Laplace neural operator}
\label{subsec:laplace_neural_operator}

To implement the proposed LNO, the input function $\mathbf f(t) \in\mathbb{R}_{d_x}$ is first transformed into a higher dimensional representation $v(t)\in\mathbb{R}_{d_z}$ using a lifting layer, denoted by $\mathcal{P}$. This layer is typically realized by a shallow neural network or a linear transformation. Next, a nonlinear operator is applied to this representation using the sum of the Laplace layer and a bias function in a neural network architecture as:
\begin{align}\label{lap_layer}
\mathbf u(t)=\sigma((\kappa(\mathbf f;\phi)*v)(t) + \mathbf W v(t)); x\in D,
\end{align}
where $\sigma$ is a nonlinear activation function, $\mathbf W$ is a linear transformation, and $\kappa$ is a kernel integral transformation. Lastly, the output $\mathbf x(t)$ is obtained by projecting $u(t)$ through a local transformation layer, $\mathcal{Q}$. In Eq.~\ref{lap_layer}, the kernel integral operator mapping is denoted as:
\begin{align}\label{lap_layer1}
(\kappa(\mathbf f;\phi)*v)(t)=\int_D \kappa_{\phi}(t,\tau,\mathbf f(t), \mathbf f(\tau);\phi)v(\tau)\;d\tau,
\end{align}
where $\kappa_{\phi}$ is a neural network parameterized by $\phi$. If we remove the dependence on the function $\mathbf f$ and impose $\kappa_{\phi}(t,\tau)=\kappa_{\phi}(t-\tau)$, Eq.~\ref{lap_layer} becomes a convolution operator:
\begin{align}\label{convolution}
(\kappa(\mathbf f;\phi)*v)(t)=\int_D\kappa_{\phi}(t-\tau)v(\tau)\;d\tau.
\end{align}
Now in the next section, we build the relationship between the Laplace transform of the input function and the Laplace transform of the output function employing the Pole-residue formulation in the Laplace layer.

\subsection{Pole-residue formulation in Laplace layer}
\label{subsec:pole_residue}

We propose to replace the kernel integral operator in Eq.~\ref{convolution} with an operator defined in the Laplace domain. By taking the Laplace transform of Eq.~\ref{convolution}, we get
\begin{align}\label{eq:linear system s domain}
U(s)= \mathscr{L}\{(\kappa(\mathbf f;\phi)*v)(t)\} = K_{\phi}(s)V(s),
\end{align}
where $ \mathscr{L}\{.\}$ and $ \mathscr{L}^{-1}\{.\}$ are operators of the Laplace transform and inverse Laplace transform, respectively; $K_{\phi}(s)= \mathscr{L}\{\kappa_{\phi}(t)\}$ and $V(s)= \mathscr{L}\{v(t)\}$.
We express $K_{\phi}(s)$ in the pole-residue form:
\begin{align}\label{eq:ht_lap}
 K_{\phi}(s)=\sum_{n=1}^{N}\frac{\beta_n}{s-\mu_n}.
\end{align}
Here, we propose to choose $K_{\phi}$ to be a neural network directly parameterized by:
\noindent
$\boldsymbol \theta = (\mu_1,\cdots, \mu_N,\beta_1,\cdots,\beta_N)$ in the Laplace domain, where $\mu_n$ and $\beta_n$ are the trainable system poles and residues, respectively. An irregular excitation signal $v(t)$ with period $T$ can be decomposed into its Fourier series:
 \begin{align}\label{eq:force-ce22}
   v(t)=\sum _{\ell=-\infty}^{\infty} \alpha_\ell\,  \exp(i\omega_\ell t),    ~~~~~~ 0 \leq t < T,
\end{align}
where $\omega_\ell=\ell \omega_1$, $\omega_1$ is the fundamental frequency in rad$/$s, and $\alpha_{\ell}$ is the complex Fourier coefficient. Taking the Laplace transform of Eq.~(\ref{eq:force-ce22}) yields:
\begin{align}\label{eq:force-pole-res2}
V(s)= \sum _{\ell=-\infty}^{\infty}  \frac{\alpha_\ell}{s-i\omega_\ell}.
\end{align}

From $U(s)=K_{\phi}(s)V(s)$, using Eqs.~\ref{eq:ht_lap} and \ref{eq:force-pole-res2}, one writes
\begin{align}\label{eq:motion_Lap1}
  U(s)= \left(\sum_{n=1}^{N}\frac{\beta_n}{s-\mu_n}\right)\left( \sum _{\ell=-\infty}^{\infty}  \frac{\alpha_\ell}{s-i\omega_\ell}\right).
\end{align}
Expressing Eq.~\ref{eq:motion_Lap1} in a pole-residue form yields~\cite{cao2023laplace,hu2016pole}
\begin{align}\label{eq:response-pol-res0}
 U(s)=\sum_{n=1}^{N}\frac{\gamma_n}{s-\mu_n}+\sum _{\ell=-\infty}^{\infty} \frac{\lambda_\ell}{s-i\omega_\ell}.
\end{align}
From Eqs.~\ref{eq:motion_Lap1} and \ref{eq:response-pol-res0}, the response residues corresponding to the first $N$ response poles (i.e., at the system poles, $\mu_n$) can be obtained by the residue theorem~\cite{kreyszig2010advanced}:
\begin{align}\label{eq:residue1}
  \gamma_n=\lim \limits_{s\rightarrow \mu_n}  (s-\mu_n) U(s)=\beta_nV(\mu_n),
\end{align}
where
\begin{align}\label{eq:vmu}
  V(\mu_n)=\sum _{\ell=-\infty}^{\infty}  \frac{\alpha_\ell}{\mu_n-i\omega_\ell}.
\end{align}
Similarly, the response residues corresponding to the last response poles (\textit{i.e.}, at the excitation poles, $i\omega_\ell$) are
\begin{align}\label{eq:residue2}
  \lambda_\ell=\lim \limits_{s\rightarrow i\omega_\ell}  (s-i\omega_\ell) U(s)=\alpha_\ell K_{\phi}(i\omega_\ell),
\end{align}
where
\begin{align}\label{eq:kphi}
   K_{\phi}(i\omega_\ell)=\sum_{n=1}^{N}\frac{\beta_n}{i\omega_\ell-\mu_n}.
\end{align}
Once $\gamma_n$ and $\lambda_\ell$ are obtained, by taking the inverse Laplace transform of Eq.~\ref{eq:response-pol-res0}, we obtain:
\begin{align}\label{eq:response-pol-res1}
 u_1(t)=\sum_{n=1}^{N}\gamma_n\exp(\mu_n t)+\sum _{\ell=-\infty}^{\infty} \lambda_\ell \exp(i\omega_\ell t).
\end{align}
In Eq.~\ref{eq:response-pol-res1}, the first summation term of the right-hand side is the transient response related to the system poles. The second summation term of the right-hand side is the familiar steady-state response operated in the frequency domain.

\noindent Next, we summarize the primary differences between FNO and LNO:
\begin{enumerate}
\item FNO chooses $K_{\phi}$ to be a neural network parameterized by $\boldsymbol \theta=(K_{\phi}(i\omega_1),\cdots, K_{\phi}(i\omega_L))$ in the frequency domain. However, LNO chooses $K_{\phi}$ to be a neural network parameterized by $\boldsymbol \theta=(\mu_1,\cdots, \mu_N,\beta_1,\cdots,\beta_N)$ in the Laplace domain, where $\mu_n$ and $\beta_n$ are the trainable system poles and residues, respectively.\\
\item FNO computes the steady-state response in the frequency domain. However, LNO computes both the transient response and the steady-state response by the Laplace domain method.
\end{enumerate}

\begin{figure}[htbp]
\centering 
\includegraphics[width=\textwidth, trim = 0cm 0cm 0cm 0cm, clip]{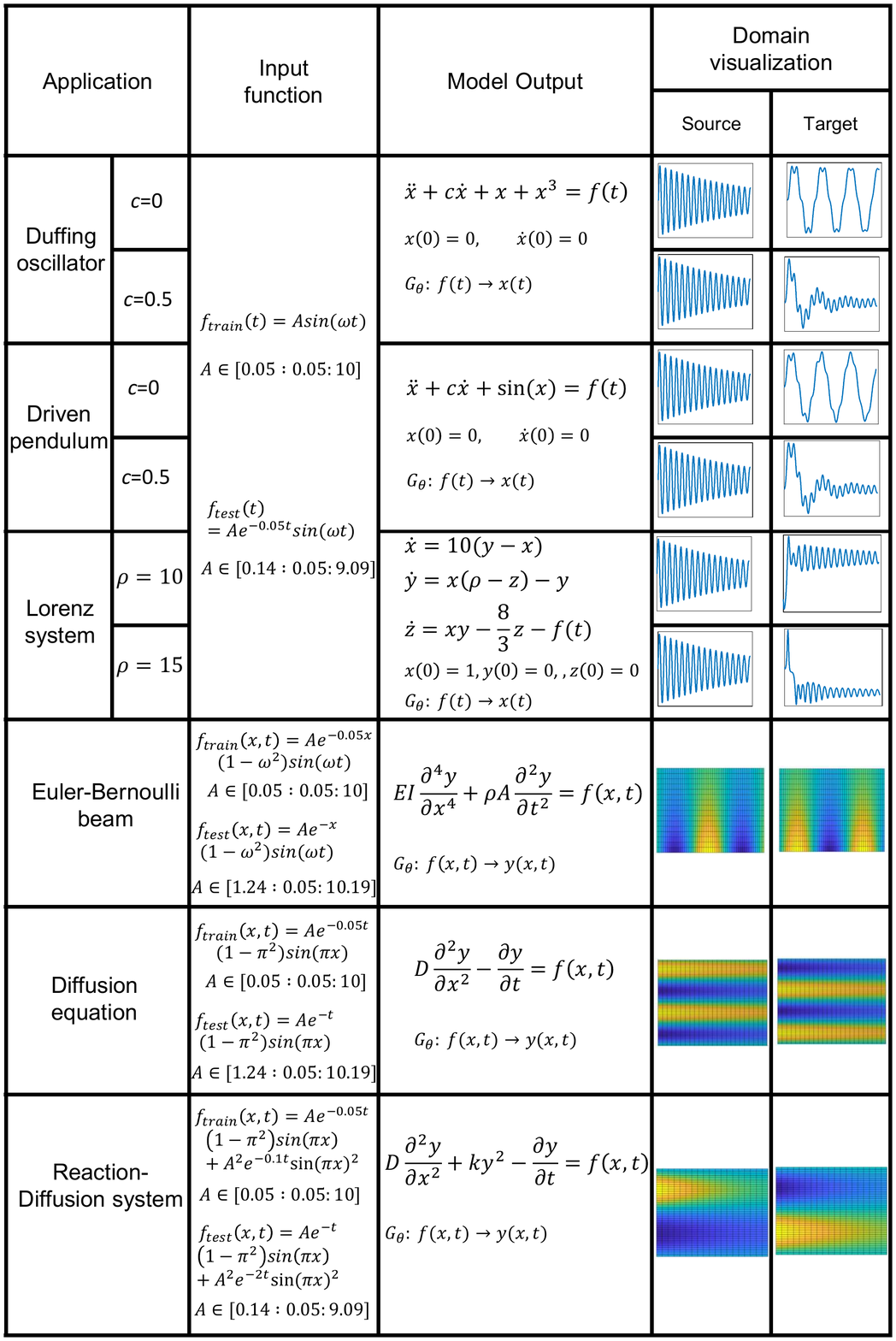}\\
\caption{A schematic representation of the examples and the subsequent experiment scenarios under consideration in this work. Shown are  representative plots of the input/output functions.}\label{summary_cases}
\end{figure}

\section{Results}
\label{sec:results}

In this section, we investigate the performance of the proposed Laplace neural operator (LNO) in comparison with the Fourier neural operator (FNO) and the gated recurrent unit (GRU), which is a promising recurrent neural network for studying three time-dependent non-linear ODEs that exhibit transient behavior and three PDEs. As GRU is suitable for sequential data, we exclude it from the PDE problems. A visual description of the different benchmarks considered in this section is presented in Fig.~\ref{summary_cases}. For each experiment, slightly different input functions are considered for the training and testing samples to investigate the generalization ability of the neural operators. The architectures of LNO, FNO, and GRU considered in each case are shown in Appendix Tables~\ref{architectures} and ~\ref{GRU_zero}. The neural operators, LNO and FNO have been implemented using the $\texttt{PyTorch}$ \cite{paszke2019pytorch}, and GRU has been implemented using $\texttt{Tensorflow}$ \cite{agarwal2016ten}. To evaluate the performance of the neural operators and GRU, we compute the relative $\mathcal L_2$ error of the predictions of the test samples and report the mean and standard deviation of this metric based on five independent training trials. A summary of the error metric of all the experiments for all five problems is shown in Fig.~\ref{L2error}.

\subsection{Duffing oscillator}
\label{subsec:duffing}

The Duffing equation is a non-linear second-order differential equation used to model certain damped and driven oscillators. The Duffing oscillator is an example of a forced oscillator with nonlinear elasticity, and can be written as:
\begin{align}\label{eq:duffing}
m\ddot{x}+c\dot{x}+k_1x+k_3x^3=f(t),
\end{align}
where $f(t)$ is the externally applied force, $m$, $c$ are the mass, damping coefficients, $k_1$ and $k_3$ stiffness of the system, and $x(t)$, $\dot{x}(t)$ and $\ddot{x}(t)$ are the displacement, velocity, and acceleration of the dynamic response, respectively. Eq.~\ref{eq:duffing} is subjected to zero initial conditions $x(0)=0$ and $\dot{x}(0)=0$, and the damping coefficient, $c\geq 0$. The Duffing oscillator is one of the prototype systems of nonlinear dynamics, which has been successfully utilized to model many processes, such as beam buckling, stiffening springs, and ionization waves in plasmas. 

In this example, for simplicity we have considered $m=1$, $k_1=1$, and $k_3=1$, and have studied two scenarios:
\begin{itemize}
\item Scenario $1$: A Duffing oscillator that does not include damping, \textit{i.e.}, $c=0$.
\item Scenario $2$: A Duffing oscillator that includes a damping term with the coefficient $c=0.5$.
\end{itemize} 

Here, our goal is to learn the non-linear operator of the system in Eq.~\ref{eq:duffing}, which maps the forcing function, $f(t)$ (considered as input) to the system response denoted by $x(t)$; that is, $\mathcal G_{\bm \theta}: f(t)\rightarrow x(t)$. To generate $N_{train}$ samples to train LNO, we consider a sinusoidal forcing function, $f_{train}(t)=Asin(5t)$, where the amplitude $A \in [0.05, 10]$ with an interval $\delta A = 0.05$, therefore $N_{train}=200$. Each sample is discretized into $2048$ temporal points and the time interval is $\Delta t = 0.01$ seconds. The response is calculated by a versatile ODE solver---$\mathbf{ode45}$ on MATLAB. For validating and testing the neural operator, we generated datasets considering a decaying sinusoidal forcing function, $f_{test}(t)=Ae^{-0.05t}sin(5t)$, where the amplitude $A \in [0.14, 9.09]$ and $N_{vali}=50$ and $N_{test}=130$.

\begin{figure}[ht]
\centering 
\includegraphics[width=\textwidth, trim = 0cm 3cm 0cm 3cm, clip]{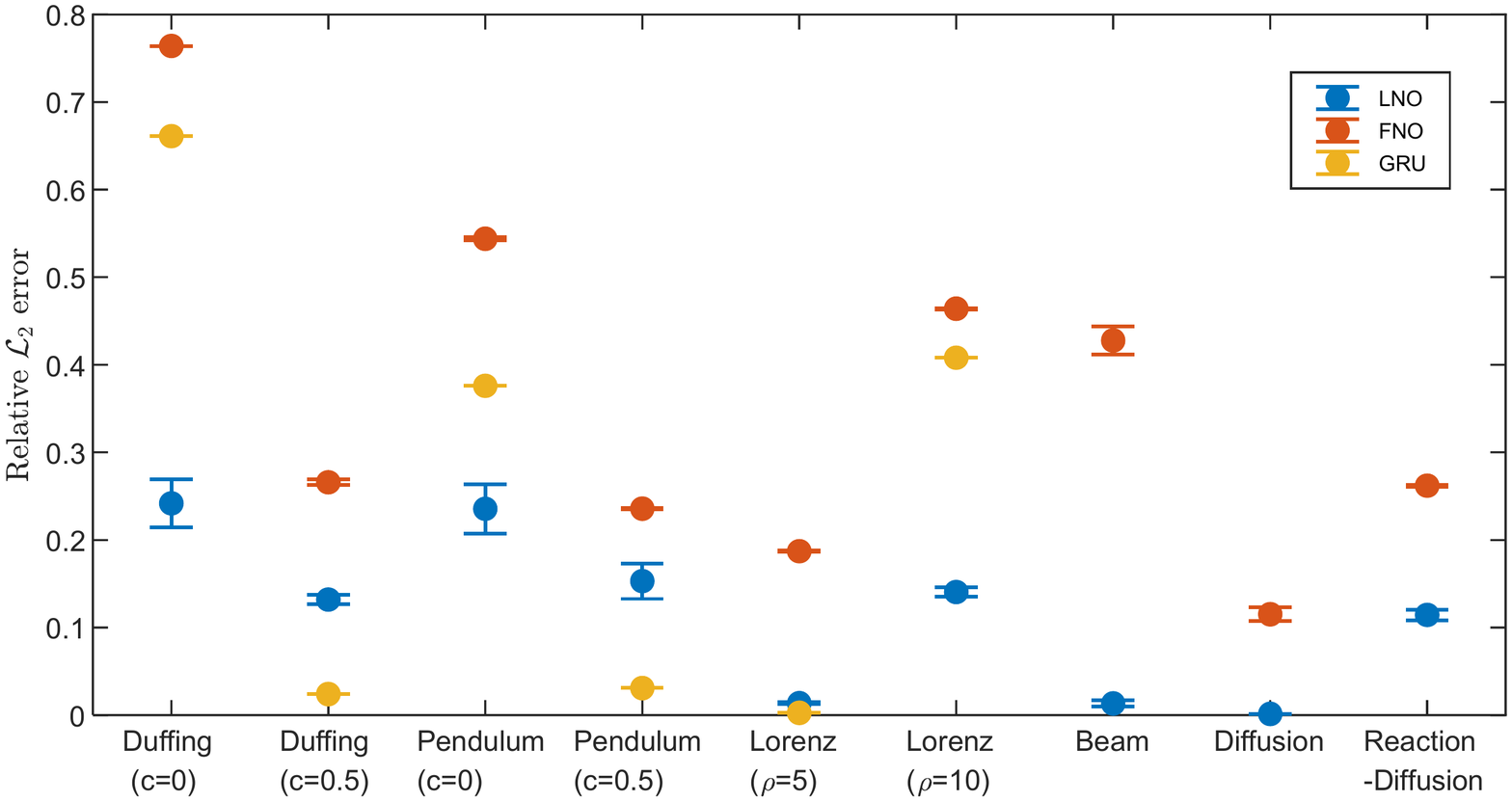}\\
\caption{Relative $\mathcal{L}_2$ error in the test cases for all the ODE and PDE cases and for different scenarios considered in each example. The plot shows the mean and the standard deviation of the error that has been computed based on five independent training trials.}\label{L2error}
\end{figure}

\begin{figure}[htbp]
\centering
\includegraphics[width=\textwidth, trim = 4cm 3cm 4cm 0cm, clip]{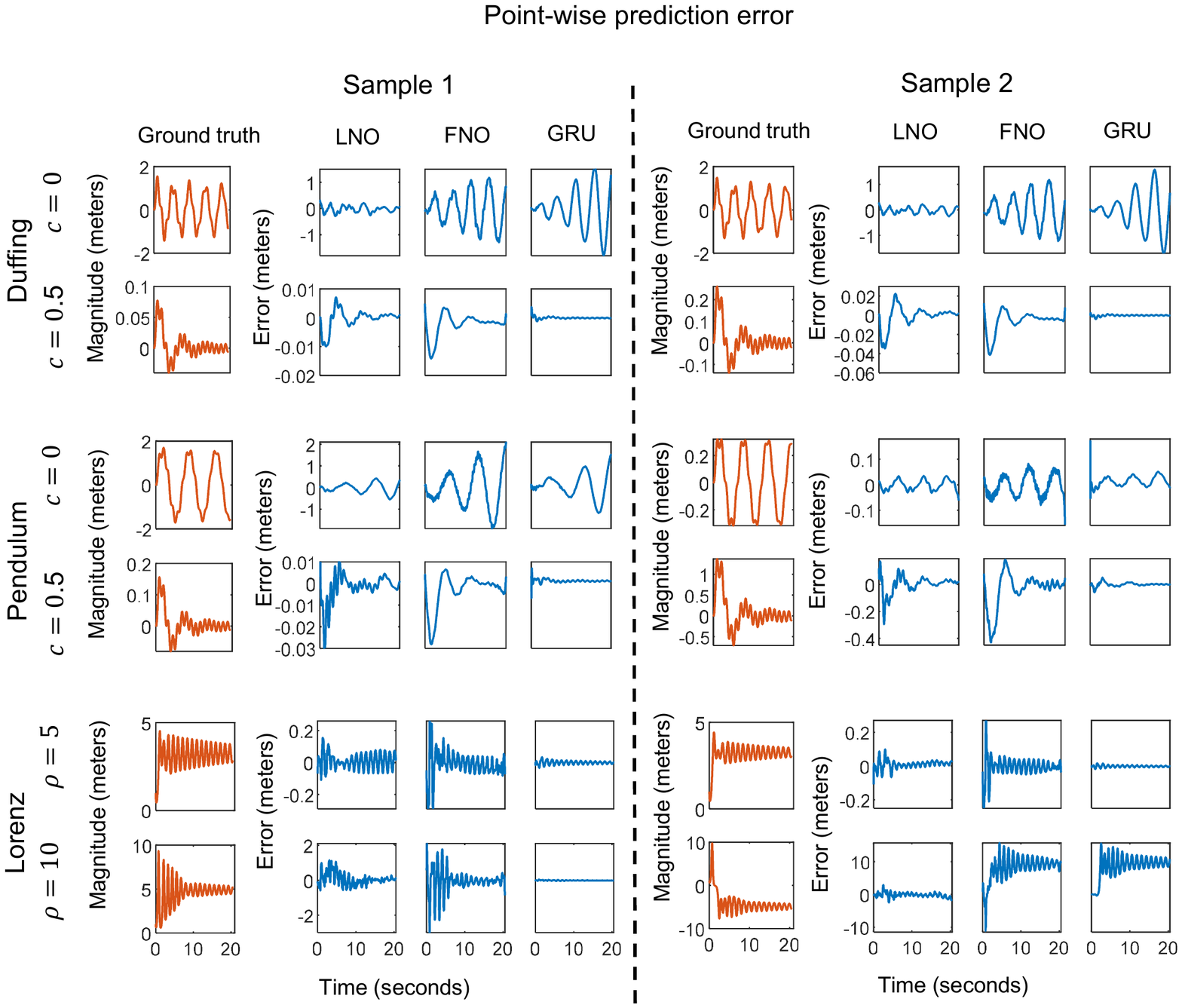}\\
\caption{Pointwise error plots of responses for two representative test samples drawn from three ODE experiments. The ground truth is plotted by red curves and the pointwise error for LNO, FNO, and GRU are presented by blue curves.}\label{samples_error}
\end{figure}

\begin{figure}[htbp]
\centering
\includegraphics[width=\textwidth, trim = 0cm 5cm 0cm 5cm, clip]{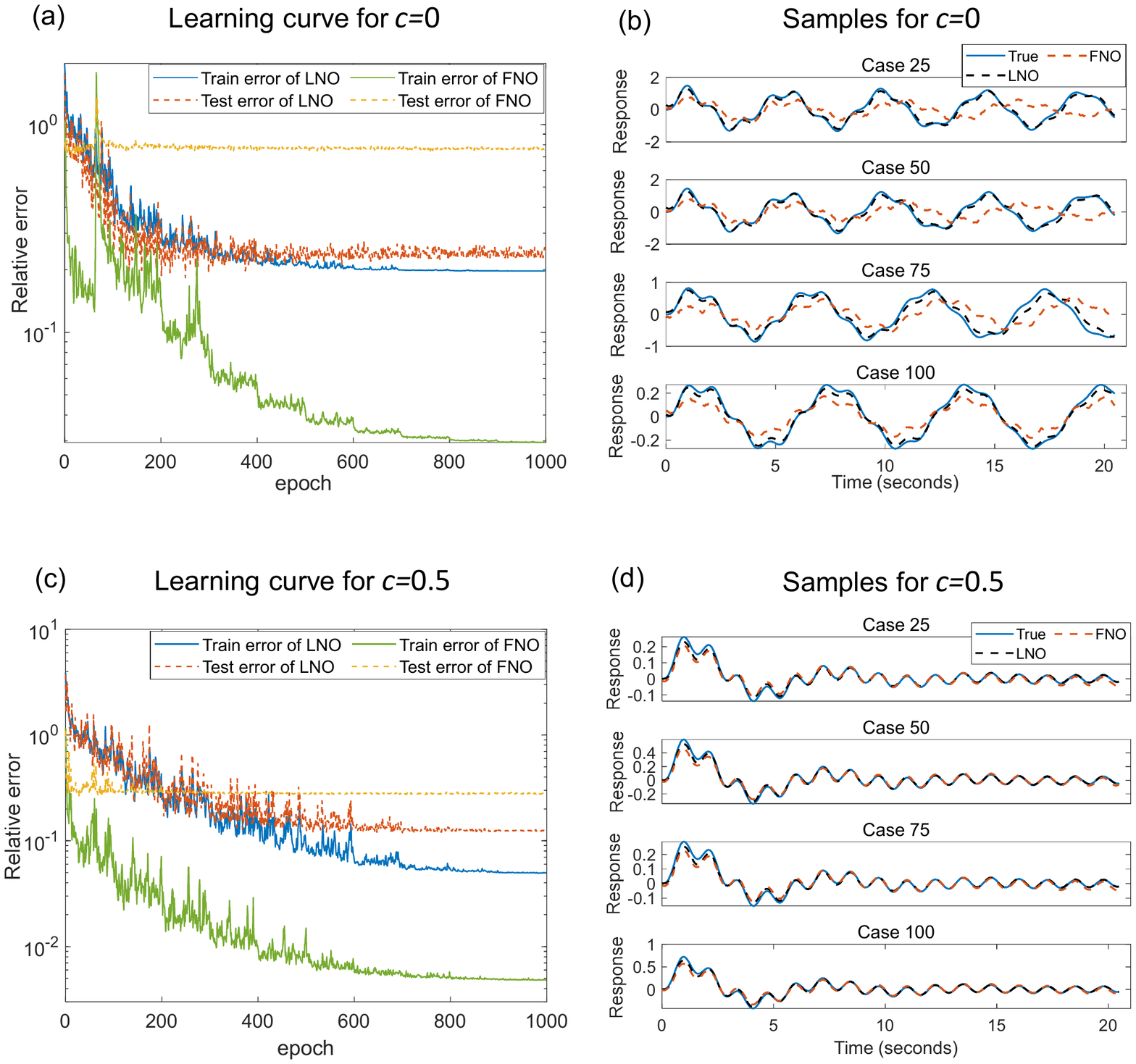}\\
\caption{Duffing oscillator: Comparison of training and testing losses and responses obtained using LNO and FNO: $(a)$ learning curve of the system without damping, $(b)$ representative response obtained from the system without damping, for test cases, $(c)$ learning curve of the system with damping $c=0.5$, $(d)$ representative response obtained from the system with damping $c=0.5$, for test cases.}\label{duffing_samples}
\end{figure}

The results presented in Fig.~\ref{L2error} demonstrate that LNO approximates the test cases of scenarios $1$ and $2$ with overall high accuracy. The improvement of the prediction of LNO over FNO is more pronounced for oscillators without damping (Scenario $1$: the first column of Fig.~\ref{L2error}). If the damping of a system is zero, there will always exist a transient response. The approximation accuracy of LNO is better than FNO because it can capture both steady-state and transient responses, although it has a higher standard deviation. LNO is more accurate than GRU when there is no damping, but GRU is more accurate than LNO when the damping term is introduced, $c=0.5$. The batch size used for LNO affects its results, so choosing an appropriate value is essential. Fig.~\ref{samples_error} (first and second rows) presents the error plots of two representative test samples for each of the neural models. Additionally, a visual representation of the prediction of the solution obtained from the two neural operators for both scenarios is shown in Fig.~\ref{duffing_samples}. Figs.~\ref{duffing_samples} (a) and (c) illustrate that the generalization gap of FNO (the difference between a model's performance on training data and its performance on unseen data) is significantly higher than that of LNO. 

\subsection{Driven gravity pendulum}
\label{subsec:example2}

In the next example, we consider a gravity pendulum subjected to an external force, $f(t)$. The equation which describes the motion of the pendulum is written as:
\begin{align}\label{eq:pendulum}
&\ddot{x}+c\dot{x}+\frac{g}{l}sin(x)=f(t),\\
\text{subject to}\;\;\;&x(0)=0, \;\; \dot{x}(0)=0,
\end{align}
where $g$ is the magnitude of the gravitational field, $l$ is the length of the rod; $c$ is the damping due to friction; $x$ is the angle from the vertical to the pendulum; and $f(t)$ is the external force. For simplicity, we have chosen $g/l=1$ in this example, and the following two scenarios are also considered:
\begin{itemize}
\item Scenario $1$: the pendulum does not include damping, \textit{i.e.}, $c=0$.
\item Scenario $2$: the pendulum includes a damping term with the coefficient $c=0.5$.
\end{itemize} 

\begin{figure}[htbp]
\centering
\includegraphics[width=\textwidth, trim = 0cm 5cm 0cm 4cm, clip]{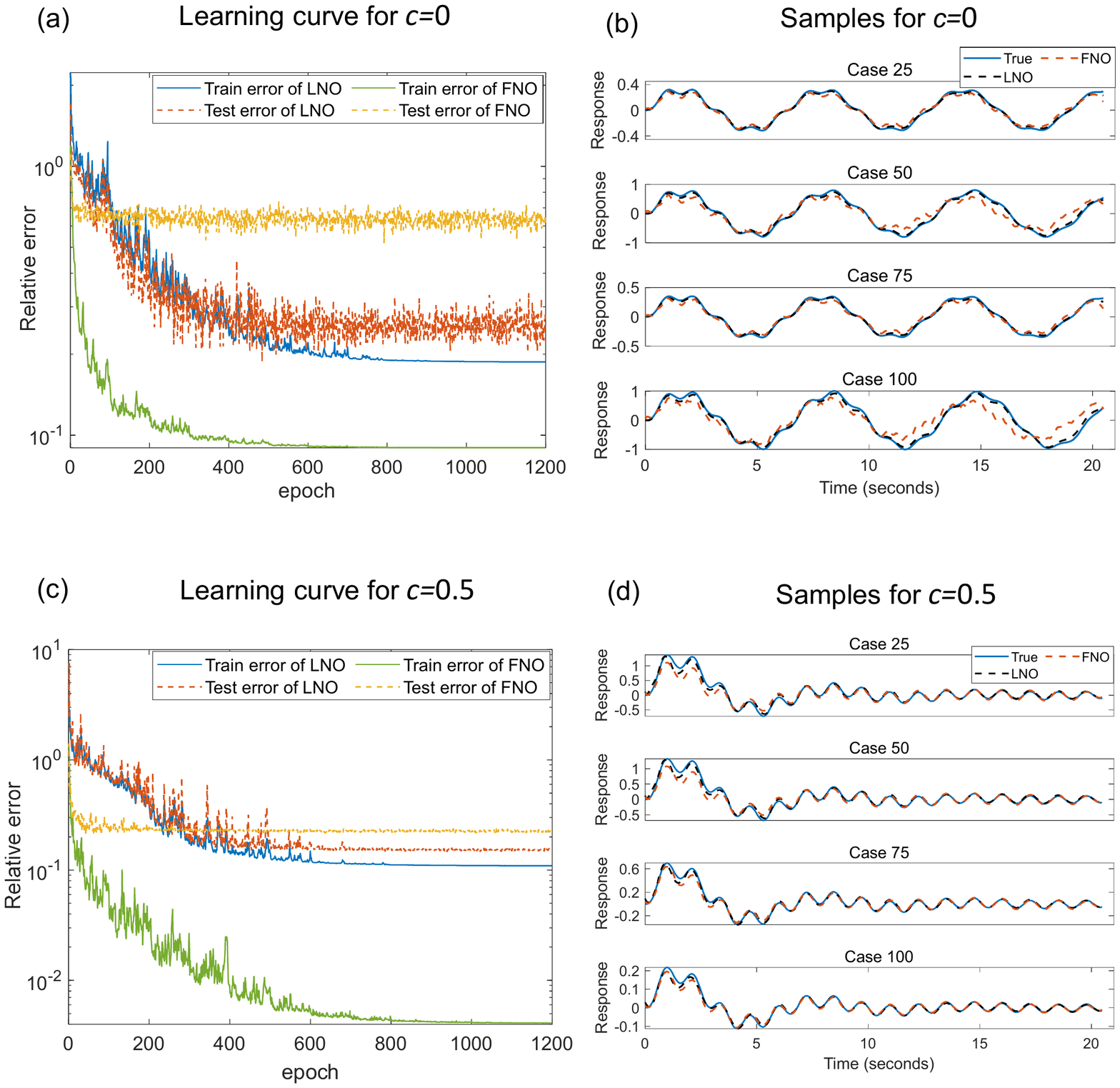}\\
\caption{Driven gravity pendulum: Comparison of training and testing losses and responses obtained using LNO and FNO: $(a)$ learning curve of the system without damping, $(b)$ representative response obtained from the system without damping, for test cases, $(c)$ learning curve of the system with damping $c=0.5$, $(d)$ representative response obtained from the system with damping $c=0.5$, for test cases.}\label{pendulum_samples}
\end{figure}

For the driven gravity pendulum model, we consider the same forcing functions for training, $f_{train}$, and testing, $f_{test}$ as used in Section~\ref{subsec:duffing}. Thus, $N_{train}=200$, $N_{vali}=50$, and $N_{test}=130$. The corresponding responses, $x(t)$ are also computed by $\mathbf{ode45}$. We aim to learn the mapping from the external force to the motion of the pendulum, that is, $G_{\boldsymbol \theta}: f(t)\rightarrow x(t)$. The third and the fourth columns of Fig.~\ref{L2error} show the relative $\mathcal{L}_2$ errors of the predictions computed by the LNO, FNO, and GRU for both scenarios. As observed in the previous example, LNO can predict the results more accurately than FNO and GRU for the systems without damping. Fig.~\ref{samples_error} (third and fourth rows) presents the error plots of two representative test samples for each of the neural models. Additionally, a visual representation of the prediction of the solution obtained from the two neural operators for both scenarios is shown in Fig.~\ref{pendulum_samples}.

\subsection{Forced Lorenz system}
\label{subsec:lorenz_system}

This example is the last ODE considered in this work. The Lorentz system is a mathematical model that simplifies many practical problems, including electric circuits, atmospheric convection, and forward osmosis. In this case, we are studying a forced Lorenz system, which has a forcing term and may be more practical. For instance, climate studies can use this system to model the temperature of the atmosphere and oceans as forcing terms~\cite{shi2021analysis}. The forced Lorenz system includes three ODEs defined as:
\begin{align}\label{eq:feedforward}
    \begin{split}
\dot{x} &= \sigma(y-x), \\
\dot{y} &= x(\rho-z)-y, \\
\dot{z} &= xy-\beta z-f(t), 
    \end{split},
\end{align}
where $x$ is proportional to the rate of convection, $y$ is proportional to the horizontal temperature variation; $z$ is proportional to the vertical temperature variation. The terms $\sigma$, $\rho$, and $\beta$ are three constant parameters related to the Prandtl number, Rayleigh number, and specific physical dimensions of the layer, respectively. Even though the equations are simple, the Lorenz system has chaotic and unpredictable behavior, so it is  highly sensitive to initial conditions. The initial conditions in this example are chosen slightly away from the state of no convection, that is $x(0)=1$, $y(0)=0$, $z(0)=0$. The $\sigma=10$ and $\beta=8/3$ are chosen in this example. We consider the following two scenarios:
\begin{itemize}
\item Scenario $1$: Rayleigh number $\rho=5$.
\item Scenario $2$: Rayleigh number $\rho=10$.
\end{itemize} 

\begin{figure}[htbp]
\centering
\includegraphics[width=\textwidth, trim = 0cm 4cm 0cm 4cm, clip]{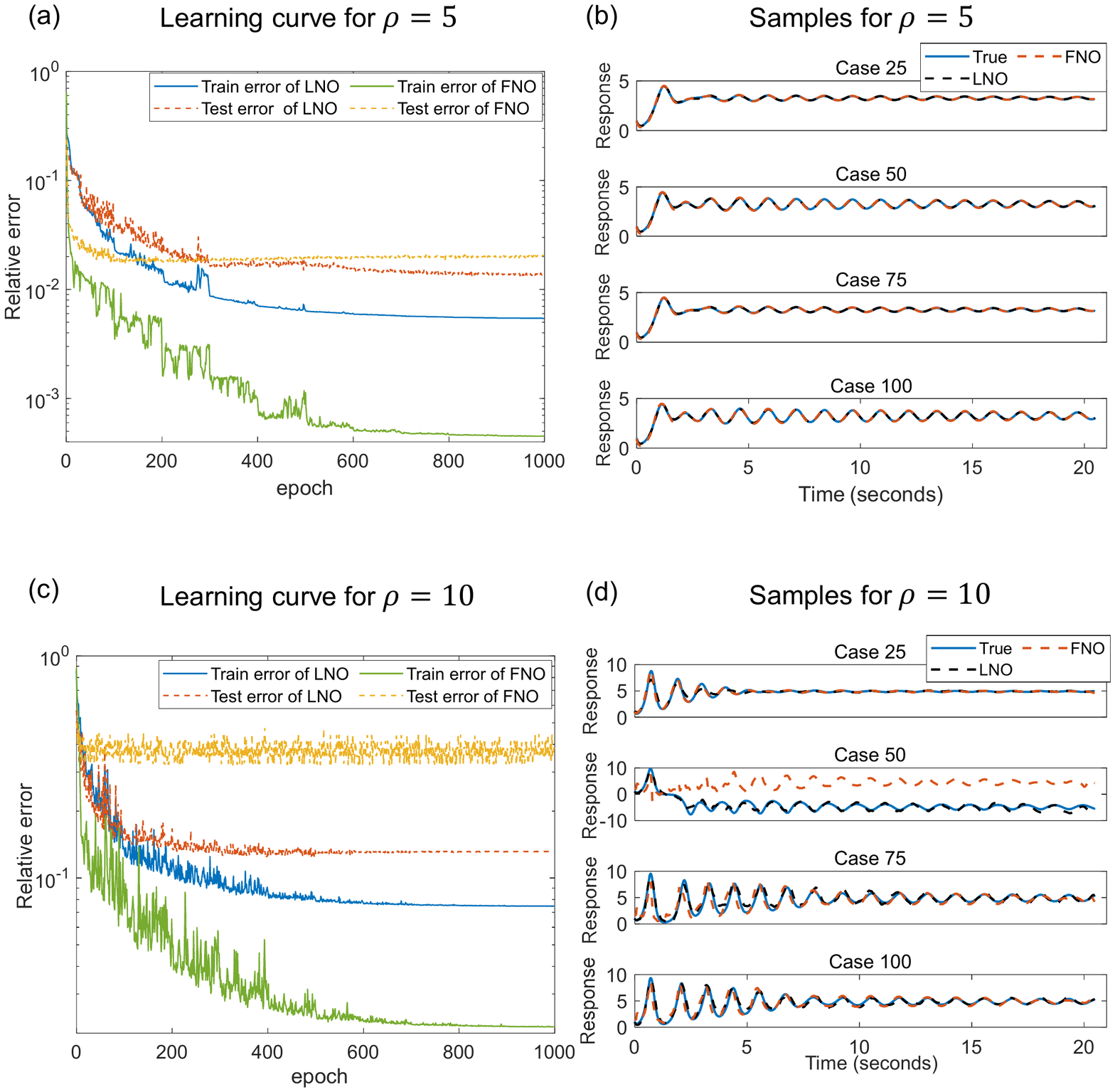}\\
\caption{Lorenz system: comparison of learning rates and responses of computed by LNO and FNO: (a) learning curve of the system with $\rho=5$, (b) response samples of the system with $\rho=5$, (c) learning curve of the system with $\rho=10$, (b) response samples of the system with damping $\rho=10$}\label{lorenz_samples}
\end{figure}

The number of samples for training, testing, and validation is kept the same as in the previous two examples. We aim to learn the mapping from the source term, $f(t)$ to the system response, $x(t)$. The results presented in Fig.~\ref{L2error} (last two columns) demonstrate that LNO approximates the test cases of scenarios $1$ and $2$ with overall high accuracy. The improvement of the accuracy is more pronounced in the case of $\rho=10$, as we observe that the response has two patterns, which are shown in the ground truth of the last row of Fig.~\ref{samples_error}. Figs.~\ref{lorenz_samples} (a) and (c) illustrate that the generalization gap of FNO is significantly higher than that of LNO. Fig.~\ref{samples_error} (last two rows) presents the error plots of two representative test samples for each of the neural models. Additionally, a visual representation of the prediction of the solution obtained from the two neural operators for both scenarios is shown in Fig.~\ref{lorenz_samples}.

\subsection{Euler-Bernoulli beam}
For a dynamic $1D$ Euler-Bernoulli beam, the Euler-Lagrange equation is written as:
\begin{align}\label{eq:beam}
 EI\frac{\partial^4y}{\partial x^4}+ \rho A\frac{\partial^2 y}{\partial t^2}=f(x,t),
\end{align}
where $y(x,t)$ is the deflection of the beam at the location $x$ and time $t$; $f(x,t)$ is the source term; $E$ and $I$ are the elastic modulus and the second moment of area of the cross-section of the beam, respectively; $\rho$ and $A$ are the material density and the area of the cross-section of the beam, respectively.

Here, our goal is to learn the operator of the system in Eq.~\ref{eq:beam}, which maps the source, $f(x,t)$, to the steady-state response $y(x,t)$; that is, $\mathcal G_{\bm \theta}: f(x,t)\rightarrow y(x,t)$. To generate $N_{train}$ samples to train LNO, we consider a function, $f_{train}(x,t)=Ae^{-0.05x}(1-10^2)sin(10 t)$, where the amplitude $A \in [0.05, 10]$ with an interval $\delta A = 0.05$, therefore $N_{train}=200$. Each sample is discretized into $51\times 17$ temporal-spatial grid points such that the time interval, $\Delta t = 0.02$ seconds, and the spatial interval,  $\Delta x = 0.1$ meters. For validating and testing the neural operator, a function, $f_{test}(x,t)=Ae^{-x}(1-10^2)sin(10 t)$, is considered, where the amplitude $A \in [1.24, 10.19]$ and $N_{vali}=50$ and $N_{test}=130$. While the analytical particular solution of Eq.~\ref{eq:beam} to $f_{train}$ is $y_{train}(x,t)=Ae^{-0.05x}sin(10 t)$, the analytical particular solution to $f_{test}$ is $y_{test}(x,t)=Ae^{-x}sin(10 t)$.

\begin{figure}[htbp]
\centering
\includegraphics[width=\textwidth, trim = 0cm 1cm 0cm 1cm, clip]{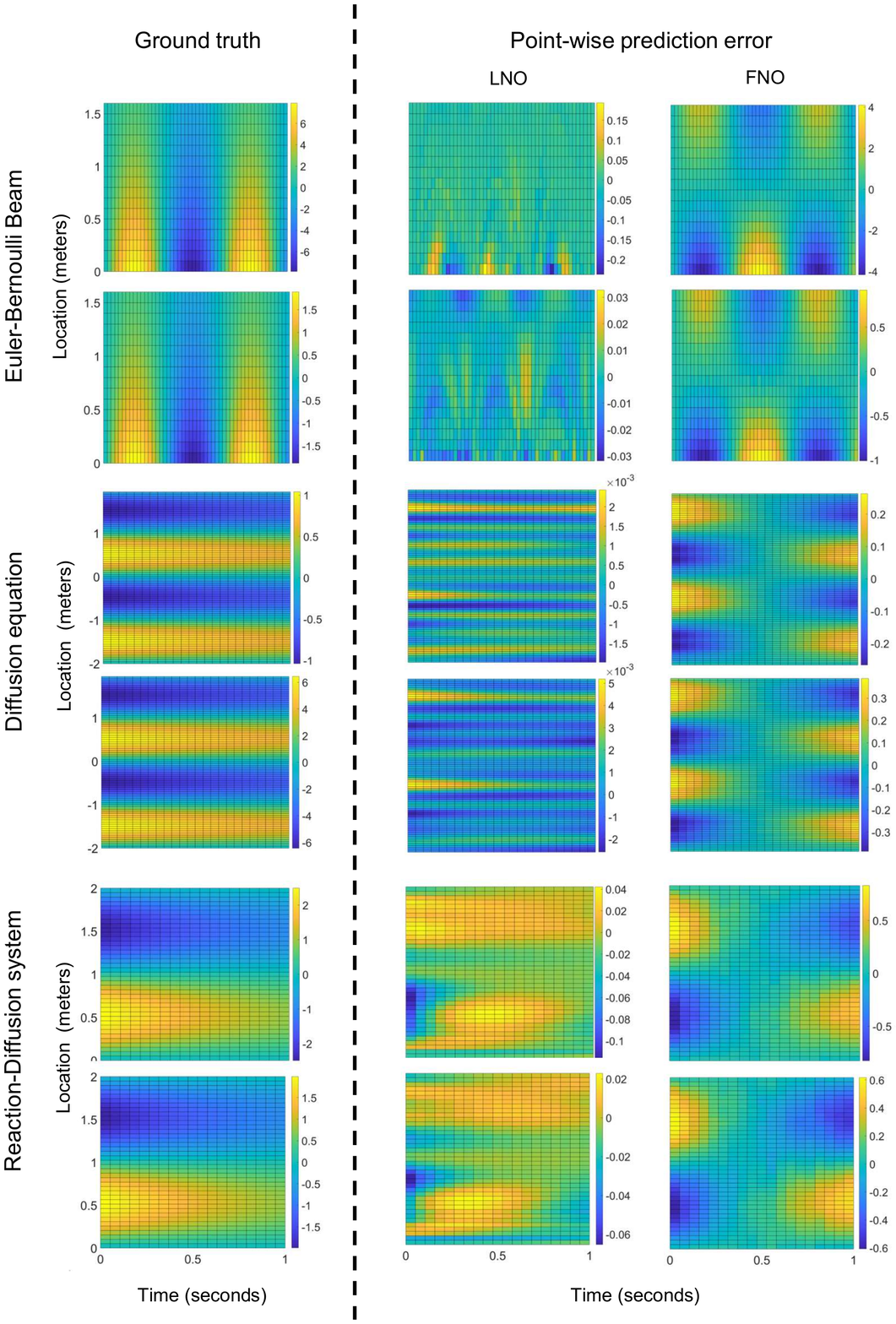}\\
\caption{Pointwise error plots of responses for two representative test samples drawn from three PDE experiments. The ground truth is plotted in the left column and the point-wise errors for LNO and FNO are presented in the right section.}\label{samples_error_pde}
\end{figure}

The third column from the end in Fig.~\ref{L2error} illustrates that the results predicted by LNO have an overall high accuracy than FNO.  Fig.~\ref{samples_error_pde} (first and second rows) presents the error plots of two representative test samples for each of the neural models. The errors of LNO are more than an order of magnitude smaller than those of FNO. 

\subsection{Diffusion equation}
\label{subsec:diffusion}
The diffusion equation is a PDE that is applied in many fields, such as information theory, material science, and biophysics. It is a special case of the convection-diffusion equation, which describes the macroscopic behavior of lots of micro-particles in Brownian motion. 
The equation is usually written as:
\begin{align}\label{eq:diffusion}
 D\frac{\partial^2y}{\partial x^2}- \frac{\partial y}{\partial t}=f(x,t),
\end{align}
where $y(x,t)$ describes the density of the diffusing material at location $x$ and time $t$; $f(x,t)$ is the source term; $D$ is the collective diffusion coefficient for density $y$ at location $x$. In this case, $D=1$ is chosen. Thus, Eq.~\ref{eq:diffusion} is identical to the heat equation.

For learning the operator of the system in Eq.~\ref{eq:diffusion}, we consider a function, $f_{train}(x,t)=Ae^{-0.05t}(1-\pi^2)sin(\pi x)$, where the amplitude $A \in [0.05, 10]$ with an interval $\delta A = 0.05$, therefore $N_{train}=200$. Each sample is discretized into $25\times 80$ temporal-spatial grid points such that the time interval, $\Delta t = 0.02$ seconds, and the spatial interval,  $\Delta x = 0.05$ meters. For validating and testing the neural operators, we generate a dataset considering the function, $f_{test}(x,t)=Ae^{-t}(1-\pi^2)sin(\pi x)$, where the amplitude $A \in [1.24, 10.19]$ and $N_{vali}=50$ and $N_{test}=130$. The analytical particular solutions of Eq.~\ref{eq:diffusion} to $f_{train}$ and $f_{test}$ are $y_{train}(x,t)=Ae^{-0.05t}sin(\pi x)$ and $y_{test}(x,t)=Ae^{-t}sin(\pi x)$, respectively.

The penultimate column in Fig.~\ref{L2error} indicates that the predictions made by LNO exhibit a higher level of accuracy than FNO overall. Fig.~\ref{samples_error_pde} (rows three and four) displays error plots for two representative test samples for each neural model. The errors of LNO are two orders of magnitude smaller than those of FNO. It is worth noting that both the Euler-Bernoulli beam and the diffusion equation are essentially learning linear operators. By utilizing the linear system to accurately represent the pole-residue formulation, the results obtained with LNO in these two cases are considerably superior to those achieved with FNO.

\subsection{Reaction-diffusion system}
\label{subsec:reaction_diffusion}
Reaction-diffusion systems describe the change in the concentration of chemical substances or particles in time and space, which can be found in chemistry, biology, geology, and physics. The diffusion-reaction equation can be represented as:
\begin{align}\label{eq:reac_diff}
D\frac{\partial^2y}{\partial x^2}+ky^2- \frac{\partial y}{\partial t}=f(x,t),
\end{align}
where $y(x,t)$ represents the concentration of chemical substances or particles at location $x$ and time $t$, $f(x,t)$ is the source term and $A$ is the amplitude of the source term. In this problem, the diffusion coefficient, $D=0.01$, and the reaction rate, $k=0.01$.

We utilize the neural operators LNO and FNO to learn the mapping from the source term, $f(x,t)$ to the steady-state response $y(x,t)$, denoted as $\mathcal G_{\bm \theta}: f(x,t)\rightarrow y(x,t)$. To generate $N_{train}$ samples for training LNO, we consider a function $f_{train}(x,t)=Ae^{-0.05t}(1-\pi^2)sin(\pi x)+A^2e^{-0.1t}sin(\pi x)^2$, where $A \in [0.05, 10]$ with an interval of $\delta A = 0.05$, resulting in $N_{train}=200$ samples. Each sample is discretized into $20\times 40$ temporal-spatial grid points with a time interval of $\Delta t = 0.0526$ seconds and a spatial interval of $\Delta x = 0.0513$. For validation and testing of the neural operator, we use a function $f_{test}(x,t)=Ae^{-t}(1-\pi^2)sin(\pi x)+A^2e^{-2t}sin(\pi x)^2$, where $A \in [0.14, 9.09]$ with $N_{vali}=50$ validation samples and $N_{test}=130$ testing samples. The analytical particular solution of $f_{train}$ for Eq.~\ref{eq:reac_diff} is $y_{train}(x,t)=Ae^{-0.05t}sin(\pi x)$, while the analytical particular solution for $f_{test}$ is $y_{test}(x,t)=Ae^{-t}sin(\pi x)$.

The final column of Fig.~\ref{L2error} shows that LNO produces more accurate results overall than FNO. Fig.~\ref{samples_error_pde} (last two rows) displays error plots of two typical test samples for each of the neural models. LNO has smaller errors than FNO, indicating that using the poles and residues of the system as network parameters aids in operator learning, even when dealing with the steady-state response of nonlinear systems.

\section{Summary}
\label{sec:summary_and_discussions}
In this work, we proposed a novel framework, called the Laplace neural operator (LNO), which parameterizes the integral kernel directly in the Laplace domain, and employs the poles and residue formulation to establish a relationship between the Laplace transforms of the functions in the input and the output spaces. The system poles and residues are the network parameters that are trained and learned in the Laplace domain, thereby making the proposed operator more interpretable. The consideration of the initial conditions and the presence of an additional exponential convergence factor in the formulation of the Laplace transformation addresses the challenges encountered by FNO when trying to approximate initial value problems, transient responses, or multiple patterns in the solution. By investigating three ODEs (Duffing oscillator, driven gravity pendulum, and Lorenz system) and three PDEs (Euler-Bernoulli beam, Diffusion equation, and Reaction-Diffusion system), we demonstrate that the new  operator, LNO, with a single module of Laplace transform predicts the response of a time-dependent system with better accuracy compared to FNO in all cases, where the FNO was architectured with four Fourier modules. Furthermore, compared to a commonly used recurrent neural network--- gated recurrent unit (GRU), LNO is more accurate in scenarios specifically considering no damping as well as systems with two or more types of patterns in the responses. From three PDE experiments, we found that expressing the neural network by trainable system poles and residues is very helpful for operator learning even though the steady-state response of nonlinear systems is considered.
Overall, LNO represents a promising new approach for learning operators that map functions between infinite-dimensional functional spaces, especially when different forms of the input functions for training and testing are considered.
\begin{itemize}
  \item Setting up different input function forms is used to investigate and demonstrate the generalization ability of LNO. If the input function forms are same, the prediction accuracy of LNO and FNO are very similar. 
  \item Our derivation of LNO begins from the convolution integral. By using the pole-residue formulation, the analytical solution is obtained. When the input $v(t)$ is the source term, the convolution integral is physically meaningful, and the relationship among the source term, system and response exactly satisfies the pole-residue formulation. However, if the input $v(t)$ is the initial conditions, the convolution integral does not have the physical meaning. Thus, the pole-residue formulation does not work significantly better than FNO. We will try to make the improvement for solving this problem.
\end{itemize}

\subsection*{\textbf{Acknowledgement}}

This work was supported by the U.S. Department of Energy, Advanced Scientific Computing Research program, under the Scalable, Efficient and Accelerated Causal Reasoning Operators, Graphs and Spikes for Earth and Embedded Systems (SEA-CROGS) project, DE- SC0023191. The authors would like to acknowledge the computing support provided by the computational resources and services at the Center for Computation and Visualization (CCV), Brown University where all experiments were carried out. 

\appendix 
\section{Network Architectures}


\begin{table}[H]
\centering
   \caption{Hyperparameters used in the LNO and FNO for training an operator to approximate the response}
   \label{architectures}
   \scriptsize
   \begin{tabular}{ccc|cccccccc}
   \hline
\multicolumn{3}{c|}{Application}  &  Layer  & Width & Mode 1 & Mode 2 & Learning rate &Batch size & Activation function & Epochs\\
\hline
\multicolumn{1}{c|}{\multirow{4}{*}{Duffing oscillator}}& \multicolumn{1}{c|}{\multirow{2}{*}{$c=0$}} & LNO &1 &4 &16 & $/$&0.002 &20 &sin &1000\\
\cline{3-11} 
\multicolumn{1}{c|}{}& \multicolumn{1}{c|}{} & FNO &4 &128 &1025& $/$ &0.002 &20 &sin &1000\\
\cline{2-11} 
 \multicolumn{1}{c|}{} & \multicolumn{1}{c|}{\multirow{2}{*}{$c=0.5$}} & LNO &1 &4 &16 & $/$&0.002 &20 &sin &1000\\
\cline{3-11} 
\multicolumn{1}{c|}{}& \multicolumn{1}{c|}{} & FNO & 4& 32& 1025& $/$&0.002 &20 &sin &1000\\
   \hline
\multicolumn{1}{c|}{\multirow{4}{*}{Driven pendulum}}& \multicolumn{1}{c|}{\multirow{2}{*}{$c=0$}} & LNO &1 &4 &20 & $/$&0.005 &40 &sin &1200\\
\cline{3-11} 
\multicolumn{1}{c|}{}& \multicolumn{1}{c|}{} & FNO &4 &32 &1025& $/$ &0.002 &40 &sin &1200\\
\cline{2-11} 
 \multicolumn{1}{c|}{} & \multicolumn{1}{c|}{\multirow{2}{*}{$c=0.5$}} & LNO & 1&4 &8& $/$ &0.002 &40 &sin &1200\\
\cline{3-11} 
\multicolumn{1}{c|}{}& \multicolumn{1}{c|}{} & FNO &4 &32 &1025 & $/$&0.002 &40 &sin &1200\\
   \hline
\multicolumn{1}{c|}{\multirow{4}{*}{Lorenz system}}& \multicolumn{1}{c|}{\multirow{2}{*}{$\rho=5$}} & LNO &1 & 4& 16& $/$& 0.005& 20&tanh &1000\\
\cline{3-11} 
\multicolumn{1}{c|}{}& \multicolumn{1}{c|}{} & FNO & 4&32 &1025& $/$ &0.002 &20 &tanh &1000\\
\cline{2-11} 
 \multicolumn{1}{c|}{} & \multicolumn{1}{c|}{\multirow{2}{*}{$\rho=10$}} & LNO & 1&4 &84 & $/$&0.002 &10 &tanh &1000\\
\cline{3-11} 
\multicolumn{1}{c|}{}& \multicolumn{1}{c|}{} & FNO &4 &32 &1025 & $/$&0.002 &20 &tanh &1000\\
   \hline
\multicolumn{1}{c|}{\multirow{2}{*}{Beam}} & \multicolumn{1}{c|}{\multirow{2}{*}{$/$}}  & LNO &1 &16 &4 &4& 0.002& 50&sin &1000\\
\cline{3-11} 
\multicolumn{1}{c|}{} & \multicolumn{1}{c|}{\multirow{2}{*}{}} & FNO & 4& 64& 9& 26& 0.002&50&sin &1000\\
   \hline
\multicolumn{1}{c|}{\multirow{2}{*}{Diffusion equation}} & \multicolumn{1}{c|}{\multirow{2}{*}{$/$}}  & LNO &1 & 16&4&4 & 0.002&50 &sin &1000\\
\cline{3-11} 
\multicolumn{1}{c|}{} & \multicolumn{1}{c|}{\multirow{2}{*}{}} & FNO &4&64 &41 &13 &0.002 &50 &sin &1000\\
   \hline
\multicolumn{1}{c|}{\multirow{2}{*}{Reaction-Diffusion system}} & \multicolumn{1}{c|}{\multirow{2}{*}{$/$}}  & LNO &1 & 48&4&4 & 0.002&50 &sin &1000\\
\cline{3-11} 
\multicolumn{1}{c|}{} & \multicolumn{1}{c|}{\multirow{2}{*}{}} & FNO &4&32 &40 &11 &0.002 &50 &sin &1000\\
   \hline
   \end{tabular}
\end{table}

\begin{table}[H]
\centering
   \caption{Hyperparameters of the GRU with zero initial conditions}
   \label{GRU_zero}
   \scriptsize
   \begin{tabular}{cc|cccccc}
   \hline
\multicolumn{2}{c|}{Application}  &  NO. of hidden layer  & Width & NO. of dense layer  & Learning rate &Batch size & Iterations\\
\hline
Duffing oscillator& $c=0$&1 &10 &1 &0.001 &128 &20000\\
Duffing oscillator& $c=0.5$&1 &10 &1 &0.001 &128 &30000\\
\hline
Pendulum& $c=0$&1 &10 &1 &0.001 &128 &20000\\
Pendulum& $c=0.5$&1 &10 &1 &0.001 &128 &30000\\
\hline
Lorenz system & $\rho=5$&1 &10 &1 &0.001 &128 &30000\\
Lorenz system & $\rho=10$&1 &20 &1 &0.001 &128 &30000\\
   \hline
   \end{tabular}
\end{table}

\bibliographystyle{elsarticle-num}
\bibliography{refer}

\end{document}